\documentclass[runningheads]{llncs}
\usepackage{graphicx}
\usepackage{xcolor}

\newcommand\blfootnote[1]{%
 \begingroup
  \renewcommand\thefootnote{}\footnote{#1}%
  \addtocounter{footnote}{-1}%
 \endgroup
}

\begin{document}
\title{Explainable Human-in-the-loop Dynamic Data-Driven Digital Twins}

\author{Nan Zhang\inst{1,2} \and
 Rami Bahsoon\inst{2} \and
Nikos Tziritas  \inst{4} \and
Georgios Theodoropoulos \inst{1,3}*
}
\authorrunning{Zhang et al.}
%
\institute{
Department of Computer Science and Engineering, Southern University of Science and Technology (SUSTech), Shenzhen, China \and
School of Computer Science, University of Birmingham, Birmingham, United Kingdom \and
Research Institute for Trustworthy Autonomous Systems, Shenzhen, China\and
Department of Informatics and Telecommunications, University of Thessaly, Greece
}
\maketitle              
\begin{abstract}
Digital Twins (DT) are essentially dynamic data-driven models that serve as real-time symbiotic ``virtual replicas'' of real-world systems. DT can leverage fundamentals of Dynamic Data-Driven Applications Systems (DDDAS) bidirectional symbiotic sensing feedback loops for its continuous updates. Sensing loops can consequently steer measurement, analysis and reconfiguration aimed at more accurate modelling and analysis in DT. The reconfiguration decisions can be autonomous or interactive, keeping human-in-the-loop. The trustworthiness of these decisions can be hindered by inadequate explainability of the rationale, and utility gained in implementing the decision for the given situation among alternatives.
Additionally, different decision-making algorithms and models have varying complexity, quality and can result in different utility gained for the model. The inadequacy of explainability can limit the extent to which humans can evaluate the decisions, often leading to updates which are unfit for the given situation, erroneous, compromising the overall accuracy of the model.       
The novel contribution of this paper is an approach to harnessing explainability in human-in-the-loop DDDAS and DT systems, leveraging bidirectional symbiotic sensing feedback. The approach utilises interpretable machine learning and goal modelling to explainability, and considers trade-off analysis of utility gained. We use examples from smart warehousing to demonstrate the approach.

\end{abstract}
\blfootnote{*Corresponding Author}

\section{Introduction}
Digital Twins (DT) are data-driven models that serve as real-time symbiotic ``virtual replicas'' of real-world systems. 
Digital Twin modelling can leverage principles of Dynamic Data-Driven Applications Systems (DDDAS) in its realisation and refinement. DT feedback loops, for example, can be designed following the DDDAS paradigm to continuously steer the measurement, modelling and analysis \cite{blasch_dddas_2018}.
As is the case in some DDDAS applications,  some DT decisions cannot be fully autonomous and need to be enacted by human operators \cite{kennedy_aimss_2007,doi:10.1260/1748-3018.5.4.561,10.1007/11758532_74}.
There can be decisions that are counter-intuitive, which may confuse the human operators. 
Humans need to understand the patterns or rationale behind the system's decision-making logic in order to trust and follow the instructions or take further decisions.
Providing explanations to humans (e.g., operators, developers or users)  can facilitate their understanding of the rationale behind the decisions and their rightfulness for a given context.
Explanations can also provide assurance for humans to trust the autonomous adaptation by DT and its underlying DDDAS. Humans are also able to examine and inspect the operation of DDDAS. In case of any anomaly or fault identified from the explanations, humans are then able to intervene in the system to assist or improve its decision-making.

The paper makes the following novel contribution:

(1) It motivates the need for providing explainability in human-in-the-loop Digital Twins, leveraging DDDAS principles and its design for feedback loops; 
(2) As a prerequisite for explanation is the identification of areas within the system, where the provision of explanation is crucial for trustworthy service; the paper describes a data-centric strategy to identify areas within DDDAS and DT that require explanations;
(3) It provides an enriched reference architecture model for DT, building on DDDAS and supported with explainability; the model extends the authors' previous work on digitally Twinning for intelligent systems and illustrates the utility gained for supporting explainability;
(4) It discusses the trade-offs and cost of providing explainability.

The rest of the paper is structured as follows.
Section \ref{Background and Related Work} provides some background on explainability.
A motivating example of smart warehouse with autonomous agents is presented in section \ref{Motivating Example}.
Section \ref{A Reference Architecture for Explainable DDDAS} presents a reference architecture of explainable Digital Twin systems, and categorises the areas in a DDDAS-based Digital Twin system where explainability can be applied.
Section \ref{Discussion} discusses the trade-off analysis of models in DTs.
Section \ref{Conclusion} concludes the paper and outlines future work.

\section{Background and Related Work}
\label{Background and Related Work}

In recent years, there has been increasing interest in the explainability and interpretability of Artificial Intelligence (AI), denoted as XAI (eXplainable Artificial Intelligence) \cite{MILLER20191}. The empirical success of deep learning models leads to the rise of opaque decision systems such as Deep Neural Networks, which are complex black-box models. Explanations supporting the output of a model are crucial in providing transparency of predictions for various AI stakeholders \cite{BARREDOARRIETA202082}.
Explanation relates to how intelligent systems explain their decisions. 
An Explanation refers to causes, and is an answer to a \textit{why-question} \cite{MILLER20191}.

In Digital Twins, decisions can be a product of autonomous reasoning processes and require explanation.
A decision model of DT may suggest locally sub-optimal decisions, but also has the risk of making mistakes.
A typical example is DTs that employ Reinforcement Learning (RL), which is a sequential decision-making model. RL aims at deciding a sequence of actions that maximises long-term cumulative rewards. However, an action that is optimal in the long term may at present be sub-optimal or even temporally lead to undesired system states. 
The lack of explanation for sub-optimal actions can lead to confusion and difficulty for humans to distinguish whether the autonomous system has made a mistake or the ``faulty'' decision is intentional for the long-term benefit.
With explanations, human operators and stakeholders can understand the rationale behind the decisions, and increase trust in the DDDAS and DT systems they utilise for decision support.

Preliminary works have been presented in discussing the explainable decision-making in dynamic data-driven Digital Twins.
In \cite{kapteyn_toward_2020}  explanations for dynamically selecting a Digital Twin model from a library of physics-based reduced-order models are provided. 
Based on sensor measurements, a trained optimal classification tree is used to match a physics-based model that represents the structural damage state at the moment of measurement. The approach is demonstrated in a case study of unmanned aerial vehicle (UAV).

The work presented in this paper differs in that it  involves the human factor in justifying explainability. Explanations can also be provided in different areas within the system architecture.

\section{Motivating Example}
\label{Motivating Example}

In this section, a motivating example based on our previous work in \cite{Zhang2020,zhang_knowledge_2022} is presented. 
The example assumes a Digital Twin for an intelligent system.
The intelligent system is composed of multiple collaborating computing nodes, each of which is controlled by an onboard intelligent agent.
Each agent learns by accumulating knowledge of its environment and interaction with other agents.
Each agent acts autonomously based on its knowledge, but its quality of decision-making may be restricted by the processing capability of the node or the power consumption if it is battery-powered. 
Therefore, the agents can act autonomously most of the time, but may need a Digital Twin of the entire system of agents for global optimisation of trade-off analysis between different goals and requirements. The resultant control decisions can be in the form of changing the decision algorithms/models of the agents \cite{zhang_knowledge_2022}, or modifying the parameters \cite{MCCUNE20141171}.
Smart warehouse is an application domain where the above-mentioned system can be applied. Multiple Autonomous Mobile Robots (AMRs) can travel within the warehouse for collaborative picking-up and delivery tasks. Each AMR is controlled by an agent that decides which task to take and which other AMRs to collaborate with. Each agent uses rules to decide its actions, but the optimality of rules can be contextual. Rules need to be adapted in order to reach optimal performance in different runtime contexts.
However, it is risky for agents to self-adapt the rules, since each agent only has partial knowledge about a subset of the entire environment in space and time. The quality of  self-adaptation is threatened by the lack of knowledge of other parts of the environment. A Digital Twin of the warehouse is able to provide more informed decisions on modifying the rules of agents, with a global view of the entire system and what-if predictions of various possible scenarios.

The Digital Twin utilises the DDDAS feedback loop to dynamically ensure both the state and knowledge of the intelligent system are accurately modelled \cite{zhang_knowledge_2022}. Based on the model, predictions of various what-if scenarios can then further inform decision-making to optimise the behaviour of the system.
Due to the nature of autonomy and high complexity within the system, decisions made by the DT may still be imperfect or even faulty. Human supervision and intervention are necessary, which requires explanations from the DT for humans to identify potentially undesired decisions.

\section{A Reference Architecture for Explainable Digital Twins Leveraging DDDAS Principles}
\label{A Reference Architecture for Explainable DDDAS}

\begin{figure}[t]
\includegraphics[width=\textwidth]{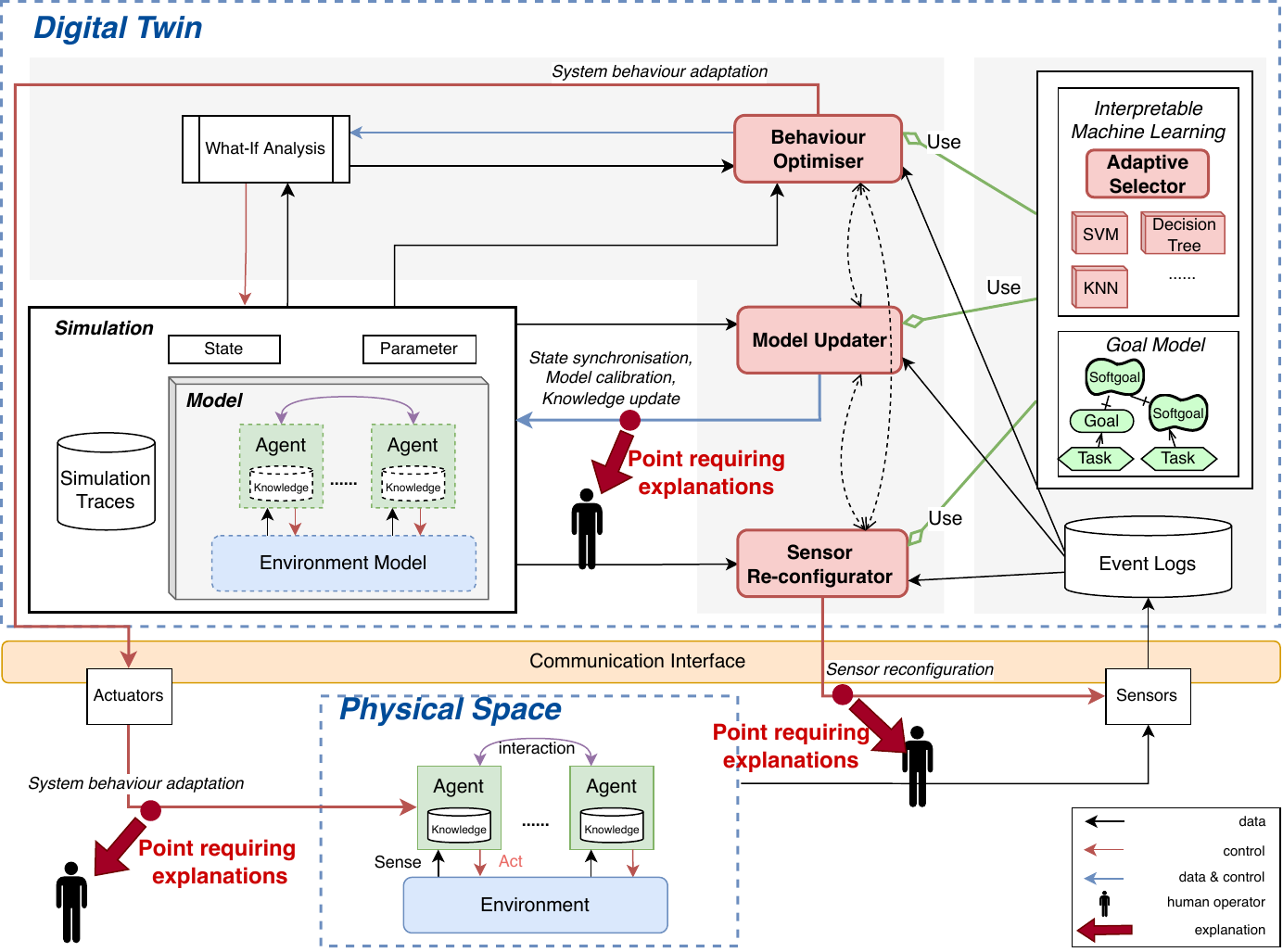}
\caption{A novel reference architecture for explainable human-in-the-loop DDDAS-inspired Digital Twins} \label{fig_arch}
\end{figure}

To address the issue of explainability in the motivating example, we present a novel 
reference architecture for an explainable  Digital Twin system, leveraging DDDAS feedback loops in Fig. \ref{fig_arch}. The architecture extends on our previous work \cite{zhang_knowledge_2022} to provide refinements of the architecture informed by DDDAS feedback loop design and enrichment with primitives for data-driven explainability. The architecture utilises interpretable machine learning and goal-oriented requirement modelling to provide explanations of decisions that are driven or influenced by data. Our architecture adopts the classical design of DDDAS computational feedback loops: (1) \textit{sensor reconfiguration}, which guides the sensor to enhance the information content of the collected data, and (2) \textit{data assimilation}, which uses the sensor data error to improve the accuracy of the simulation model \cite{blasch_dddas_2018}. Additionally, we refine the above two loops with human-in-the-loop inputs, as explainability is essentially a human-centric concern.        Utilising the simulation model and analysis, the system behaviour can be dynamically controlled in a feedback manner.

In Fig. \ref{fig_arch}, the physical space at the bottom contains the system that is composed of intelligent agents, as in the motivating example. The Digital Twin contains a simulation model that replicates the physical space. The agents in the model are identical software programs to the agents in the physical space.

There are three controllers that utilise feedback loops for autonomous decision-making: \textit{Sensor Re-configurator}, \textit{Model Updater}, and \textit{Behaviour Optimiser}.
Each controller incorporates interpretable machine learning and goal models to provide explanations to human operators.
The three control components can also interact with each other. For instance, the \textit{Behaviour Optimiser} may lack high fidelity data to justify and validate its decision, then it can notify \textit{Sensor Re-configurator} asking for further enhanced sensory data.

\subsection{Explainable Decisions for Human-in-the-loop Digital Twins}
\label{Explainable Decisions for Human-in-the-loop}

Human factors can be present in the two computational feedback loops in DDDAS-based DT and can be captured in an additional physical human-in-the-loop feedback process, as illustrated in Fig. \ref{fig_arch}. 
Explanations are needed to enhance the trust of humans in the system and/or allow humans to find anomalies or faults in the autonomous process in order to improve the decisions made by the system.
With explanations provided, humans can trace the input data and how the data are assimilated to generate the adaptation decision. Inadequate decisions can be identified and their cause can be located: because of faulty input data or during the reasoning process.

Explanations are closely linked to decisions that are steered by data, accumulated knowledge or computations by data assimilated into the system. Our architecture aims at providing explanations on how and why  decisions or adaptations are made, linked to data: the data that support a decision need to be traced, since decision is a result of data input and how the data is then processed and/or manipulated.
In the architecture, the adaptations of the three controllers represent three categories of decisions that can be made by the DDDAS-based Digital Twin: (1) \textit{measurement adaptation}, (2) \textit{model adaptation}, and (3) \textit{system behaviour adaptation}.
For each type of adaptation decision, explanations are required to increase the transparency of decisions for humans (shown as ``Point requiring explanations'' in Fig. \ref{fig_arch}).
Goal models help in presenting the satisfaction of system requirements by their utility. Each controller uses interpretable machine learning for its decision-making to directly provide explanations.

\subsubsection{Explanation for Measurement Adaptation}

Measurement adaptation is a part of the sensor reconfiguration feedback loop, which controls what and how data are collected. This type of adaptation is decided by the \textit{Sensor Re-configurator} in the architecture.
Explanations are essential to answer: what is the purpose of a measurement adaptation, and how is it beneficial?
Examples of reasons can be to more accurately estimate the state, system behaviour adaptation requiring more data for validation, or cost-effective concerns such as energy.

Different aspects of adaptation on the measurement are as follows. Each adaptation requires explanations.
\begin{itemize}
\item \textbf{Data source}: The Digital Twin can dynamically decide what sources of data to be collected. In the motivating example, knowledge and state are two sources of data. When identifying discrepancies between the model and the real system, comparing knowledge between the two can lead to more cost-efficient results than comparing states in certain scenarios \cite{zhang_knowledge_2022}.
\item \textbf{Sampling rate}: The frequency of sensing can be adjusted. A sensor can be put into sleep mode for the reason of energy saving, if its readings can be predicted by historical data or other data sources \cite{Minerva_iot_2020}.
\item \textbf{Sensing fidelity}: Sensors can send back imagery data in high or low resolution, based on the trade-off between data enhancement due to higher fidelity observation vs. cost \cite{blasch_dddas_2018}.
\item \textbf{Sensor placement}: The locations of sensors can be adjusted to monitor areas of interest possibly with multiple sensors from different perspectives.
\item \textbf{Human monitoring}: Humans can follow instructions from the Digital Twin to monitor and inspect the system and environment on the spot and report data back. 
\end{itemize}

\subsubsection{Explanation for Model Adaptation}

The aim of model adaptation is to increase the fidelity of the simulation model. This adaptation decision is made by the \textit{Model Updater} in Fig. \ref{fig_arch}.
Explanations should justify why the adaptations are beneficial in improving fidelity.
The adaptation can be state replication and estimation, parameter calibration, and knowledge update.

Data assimilation is a common approach for state estimation, which aims at finding a ``best'' estimate of the (non-observable) state of the system.
However, the motivating example is different from the conventional application scenarios of data assimilation such as Kalman Filter. 
The system is discrete-event and involves complex interactions between different entities, which cannot be easily characterised by mathematical properties and assumptions with numerical models. 
Although a data assimilation framework for discrete event simulation models has been proposed \cite{Hu_assimilation_DES_2019},
the example system itself is driven by inherent computation and knowledge of the agents, which poses an additional challenge for the model to replicate not only the state but also the knowledge of the system.

There have already been some research efforts in providing explanations for model adaptation. Interpretable machine learning can be used to select within a library of models based on real-time observation \cite{kapteyn_toward_2020}.
Model adaptation can be driven by the discrepancies between sensor data and the data from the Digital Twin model. In \cite{gao_anomaly_2021}, a Gaussian Mixture Model-based discrepancy detector is first designed to identify anomalies. Then the detected anomalies are further classified into different types by Hidden Markov Model. The classification of anomalies can serve as the explanation for discrepancies to support later decisions. However, their work does not involve feedback loops that correct the anomalies after classification.

\subsubsection{Explanation for System Behaviour Adaptation}

System behaviour adaptation refers to how the Digital Twin model is used for simulating different what-if scenarios to inform decisions on optimising the system behaviour. This type of decision is made by the \textit{Behaviour Optimiser} in Fig. \ref{fig_arch}.

What-if analysis can be self-explainable, 
since \textit{what if}-questions are essentially asking explanatory questions of \textit{how}, and are just a contrast case analysing what would happen under a different situation or with an alternative choice \cite{MILLER20191}.
Nevertheless, the process of exploring all different what-if scenarios can be expensive thus requiring to be optimised by e.g. searching algorithms.

Explanations can also be achieved by providing the satisfaction of the system requirements during operation \cite{Welsh2014}. These requirements are imposed by experts during the design time of the Digital Twin system, which can involve the task goals (e.g. delivery efficiency in the warehouse example) or system performance (e.g. computational load of the Digital Twin). Goal models can be utilised to provide explanations \cite{Welsh2014}, which is also incorporated in our architecture.

\section{Discussion}
\label{Discussion}
\subsection{Trade-off Analysis}
Our architecture has taken a data-driven stance to explainability of decisions in the system. However, dynamism in data sensing, assimilation and processing can make unnecessary explainability costly.  
Although explanation is beneficial for transparency, trustworthiness and understandability, the trade-off between performance and explainability needs to be considered in dynamic and data-driven systems. Additionally, decision-making algorithms and models differ in quality of decisions; henceforth, the levels of explanation needed \cite{BARREDOARRIETA202082} can vary with the quality of decisions induced by these models. The empirical evidence that evaluates how end users perceive the explainability of AI algorithms has been established \cite{herm_stop_2022}. The trade-off is situational and depends on the complexity of the problem. Decision models may rank differently in performance vs. explainability when solving different problems, along with the varying complexity of input and training data (e.g. tabular data or images) \cite{herm_stop_2022}.
For highly dynamic data-driven open systems, the data sensing can vary in complexity, quantity and refresh rate. The type and sophistication of the decisions can also vary.
There is a need for DT that self-adaptively decides on when to use which decision models and the level of explanation required. 
In the reference architecture in Fig. \ref{fig_arch}, an \textit{Adaptive Selector} is proposed to dynamically switch between the models in different runtime situations.
One possible solution to adaptive selection is to abstract the system into different states. Each state can be pre-calculated for its requirement on the level of explanation. The \textit{Adaptive Selector} monitors the state to make selections.

The choice of decision models may further influence the trust of human operators and their willingness to enact the decisions. DT's dynamic trade-off analysis demands awareness of the cost due to the explanation's impact on humans' behaviour.
An adaptive cost-aware approach has been proposed in \cite{li_reasoning_2020}, which uses probabilistic model checking to reason about when to provide explanations in a case study of web services. This study assumes that explanation can improve humans' probability of successfully performing a task, but will incur delayed human actions, because humans need time to comprehend the explanation.

\subsection{Architecture Evaluation}


The inclusion of explainability is viewed as an architecture design decision. Well-established architectural evaluation methods may be used to evaluate the extent to which explainability can improve the utility of the architecture under design, against other quality attributes of interest that need to be satisfied within the architecture.

Scenario-based evaluation methods, such as ATAM \cite{KazmanATAMMethod2000} may be used during the inception and development stages of the architecture to inform the design of explainability.
Scenarios are human-centric and context-dependent. Types of scenarios can be typical, exploratory, or stress scenarios. The extent to which the design for explainability is fit depends on the choice of scenarios. Scenarios may focus on issues such as the need, effectiveness and added value of having explanation embedded as part of the architecture.
Examples of the added value of supporting explainability in the data-driven cycle include: ensuring compliance (safety, security, etc.) and avoiding accidental violation of compliance as data is streamed, analysed, processed and/or decisions are actuated; 
tracing back the cause of issues attributed to data and decisions, mainly for debugging and forensic investigation; transparency of decisions and control by explaining data that can drive control; 
explaining the risk of critical data-driven decisions, among the many others.

Considering the \textit{Updater} as an example, the parameter calibration can be triggered by policies and the measured discrepancy between the simulation and the real system. Calibration should be triggered only when necessary, since estimating the parameters requires excessive computational resources. However, issues such as sensor anomaly may trigger the calibration by mistake, causing unnecessary computational costs. Explanations can inform humans to trace the need for calibration and identify such a measurement anomaly. Frequent explanations contribute to cautious anomaly identification by always involving humans, but can instead cause annoyance and decreased quality of experience (QoE) of the humans when even trivial decisions need confirmation from humans. The trade-off analysis between computational cost and QoE is essential in evaluation.

\section{Conclusion}
\label{Conclusion}

This paper has aspired to investigate the problem of explainability in human-in-the-loop dynamic data-driven Digital Twin systems. We have contributed to a novel reference architecture of DT, which draws inspiration from the design of DDDAS feedback loops and enrich it with primitives for explainability.
The architecture is designed to identify and elaborate on where and when explanations can support decisions that are data-driven or influenced by data assimilated and/or computed by the system.

The areas that need explanations are identified by focusing on three types of adaptation decisions, traced to data. We have refined and enriched the architecture to provide explainability linked to the decisions of three controllers. The controllers use interpretable machine learning and goal models to provide explanations. Future work will focus on further refinements and implementation of variants of explainable Digital Twin architectures for autonomous systems. We will look at cases from smart warehouse logistics and manufacturing, where humans and machines collaborate to inform the design and evaluate the effectiveness of adaptive explanation. Special attention will be paid to the impact of explanation on humans' behaviour. The \textit{Adaptive Selector} in the architecture will be instantiated by algorithms that use state-based approaches to dynamically switch the decision-making models. The work is ongoing and a prototype  agent-based modelling testbed of the motivating example has already been implemented in AnyLogic\footnote[1]{https://www.anylogic.com/}.

\section*{Acknowledgements}

This research was supported by: Shenzhen Science and Technology Program,  China (No. GJHZ20210705141807022); SUSTech-University of Birmingham Collaborative PhD Programme; Guangdong Province Innovative and Entrepreneurial Team Programme, China (No. 2017ZT07X386); SUSTech Research Institute for Trustworthy Autonomous Systems, China; and EPSRC/EverythingConnected Network project on Novel Cognitive Digital Twins for Compliance, UK.

%
\bibliographystyle{splncs04}
\bibliography{references}

\end{document}